\documentclass[11pt]{article}
\usepackage{eacl2017}
\eaclfinalcopy

\usepackage{times}
\usepackage{latexsym}
\usepackage{url}
\usepackage{arydshln}
\usepackage{graphicx}
\usepackage{amsmath}
\usepackage{amsfonts, amssymb}
\usepackage{xcolor}

\newcommand{\todooff}{\long\gdef\todo##1{}}
\newcommand{\todoon}{\long\gdef\todo##1{{
\bf\textcolor{red} {TODO: ##1}
}}}
\todoon
\todooff

\newcounter{notecounter}
\newcommand{\enotesoff}{\long\gdef\enote##1##2{}}
\newcommand{\enoteson}{\long\gdef\enote##1##2{{
\stepcounter{notecounter}
{\large\bf
\hspace{1cm}\arabic{notecounter} $<<<$ ##1: ##2
$>>>$\hspace{1cm}}}}}
\enoteson
\enotesoff

\def\dnrm#1{\mbox{$_{\hbox{\scriptsize #1}}$}}
\def\uprm#1{\mbox{$^{\hbox{\scriptsize #1}}$}}

\def\figref#1{Figure~\ref{fig:#1}}

\def\tabref#1{Table~\ref{tab:#1}}
\def\tablabel#1{\label{tab:#1}\label{p:#1}}

\def\secref#1{Section~\ref{sec:#1}}
\def\seclabel#1{\label{sec:#1}\label{p:#1}}
\def\eqref#1{Eq.~\ref{eqn:#1}}
\def\eqlabel#1{\label{eqn:#1}}

\def\eaclpaperid{43}

\title{Multi-level Representations \\ for Fine-Grained Typing
of Knowledge Base Entities}

\author{Yadollah Yaghoobzadeh \rm{and} \textbf{Hinrich Sch{\"u}tze}\\
  Center for Information and Language Processing \\
  LMU Munich, Germany \\
  {\tt yadollah@cis.lmu.de} 
}

\date{}
\begin{document}
\maketitle

\begin{abstract}
Entities are essential elements of natural language.
In this paper, we present methods for learning multi-level
representations of entities on three complementary levels: 
\textbf{character}
(character patterns in entity names extracted, e.g., by neural
networks), \textbf{word} (embeddings of words in entity names) and
\textbf{entity} (entity embeddings). 
We 
investigate
state-of-the-art learning methods on each level and find
large differences, e.g., for
deep learning models, traditional ngram features
and the subword model of \texttt{fasttext} \cite{subword16}
on the \textbf{character} level;
for 
\texttt{word2vec}
\cite{mikolov2013efficient} on the \textbf{word} level; and
for the order-aware model \texttt{wang2vec} \cite{wang2vec15}
on the \textbf{entity} level.

We confirm experimentally that each level of representation contributes
complementary information
and a joint
representation of 
all three levels improves  the existing embedding based baseline
for fine-grained entity typing
by a large margin.
Additionally, we show that adding information from 
entity descriptions 
further improves multi-level representations
of entities.

\end{abstract}

\section{Introduction}
Knowledge about entities is essential for understanding
human language.  This knowledge can be attributional (e.g.,
canFly, isEdible), type-based (e.g., isFood, isPolitician,
isDisease) or relational (e.g, marriedTo, bornIn).
Knowledge bases (KBs) are designed to store this information
in a structured way, so that it can be queried easily.
Examples of such KBs are Freebase
\cite{bollacker2008freebase}, Wikipedia, Google knowledge
graph and YAGO \cite{suchanek2007yago}.  For automatic
updating and completing the entity knowledge, text resources
such as news, user forums, textbooks or any other data in
the form of text are important sources.  Therefore,
information extraction methods have been introduced to
extract knowledge about entities from text.  In this paper,
we focus on the extraction of entity types, i.e.,
assigning types to -- or \emph{typing} -- entities. 
Type information can help extraction of relations by 
applying constraints on relation arguments.

We address a problem setting in which the following  are
given: a KB with a set of entities $E$, a set of types $T$
and 
a membership function 
$m:  E \times T \mapsto \{0,1\}$ such that
$m(e,t)=1$ iff entity $e$ has type $t$;
and a large corpus $C$ in which mentions of 
$E$ are annotated. 
In this  setting, we address the task of
\emph{fine-grained entity typing}: we want to
learn a probability function $S(e,t)$ for a pair of entity $e$ and type $t$
and based on $S(e,t)$
infer whether $m(e,t)=1$ holds, i.e., whether entity $e$ is a
member of type $t$. 

We address this problem by learning a multi-level
representation for an entity that contains the information
necessary for typing it.  One important source is the
\emph{contexts in which the entity is used}.  We can take
the standard method of learning embeddings for words and
extend it to learning embeddings for entities. This requires
the use of an entity linker and can be implemented by
replacing all occurrences of the entity by a unique token.
We refer to entity embeddings as \emph{entity-level
  representations}.  Previously, entity embeddings have been
learned mostly using bag-of-word models like
\texttt{word2vec} (e.g., by \newcite{Wang14joint} and
\newcite{yyhs15fig}).  
We show below that order information
is critical for high-quality entity embeddings.

Entity-level representations are often uninformative for
rare entities, so that using only entity embeddings
is likely to produce poor results. 
In this paper, we use
\emph{entity names}
as a source of information that is
complementary to entity embeddings. 
We define an entity name as a 
noun phrase that is used to refer to an entity.
We learn character and word level representations of 
entity names.

For the \emph{character-level representation}, 
we adopt different character-level 
neural
network architectures. 
Our intuition is that there is
sub/cross word information, e.g., orthographic patterns, that is
helpful to get better entity representations, especially for
rare entities. A simple example is that a three-token
sequence containing an initial like ``P.'' surrounded by two
capitalized words (``Rolph P.\ Kugl'') is likely to refer to a person.

We compute
the \emph{word-level representation} as the sum of the
embeddings of the words that make up the entity name.
The sum of
the embeddings accumulates evidence for a
type/property over all constituents, e.g., 
a name
containing  ``stadium'', ``lake'' or
``cemetery'' is likely to refer to a location. 
In this paper, we compute our word level representation 
with two types of word embeddings:
(i) using only contextual information of words in the corpus, e.g., 
by \texttt{word2vec} \cite{mikolov2013efficient}
and
(ii) using subword as well as contextual information of words,
e.g., by Facebook's recently released  \texttt{fasttext} \cite{subword16}. 

In this paper, we integrate 
character-level and word-level with entity-level representations
to
improve the results of previous work on fine-grained typing
of KB entities. 
We also show how descriptions of entities in a KB 
can be a complementary source of information to our 
multi-level representation to improve 
the results of entity typing, especially for rare entities.

Our main contributions in this paper are:
\begin{itemize}

\item 
We propose new methods for learning entity representations 
on
three levels: character-level, word-level
and entity-level.

\item 
We show that these levels are complementary
and a joint model that uses all three levels
  improves the state of the art on the task of fine-grained
  entity typing by a large margin.

\item We experimentally show that an order dependent embedding 
is more informative than its bag-of-word counterpart for entity representation. 
\end{itemize}
We release our dataset and source codes:
\url{cistern.cis.lmu.de/figment2/}.

\section{Related Work}
\textbf{Entity representation.}
Two main sources of information used for learning 
entity representation are: 
(i) links and descriptions in KB,
(ii) name and contexts in corpora. We focus
on name and contexts in corpora,
but we also include
(Wikipedia) \enote{hs}{last word: check} descriptions.
We represent entities on three levels: entity, word and character.
Our  entity-level representation is similar to work 
on relation extraction \cite{Wang14joint,wang16ijcai}, 
entity linking \cite{yamada16linking,fang2016coNLL},
and entity typing
\cite{yyhs15fig}.
Our word-level representation with distributional word embeddings
is similarly used 
to represent entities for entity linking \cite{sun15entity}
and relation extraction
\cite{socher2013reasoning,Wang14joint}.
Novel entity representation methods we introduce in this paper are
representation based on
\texttt{fasttext} \cite{subword16} subword embeddings,
several character-level representations,
``order-aware'' entity-level embeddings and
the combination of several different representations into one multi-level representation.

\textbf{Character-subword level neural networks.}
Character-level convolutional
neural networks (CNNs) 
are applied by
\newcite{Santos14pos}
to part of speech (POS) tagging,
by \newcite{Santos15ner}, \newcite{ma2016}, 
and \newcite{chiu2016} to named
entity recognition (NER), by 
\newcite{Zhang15ch} and
\newcite{Zhang15scratch} to sentiment analysis and text categorization,
and by 
\newcite{kim15} to language modeling (LM).
Character-level LSTM is applied by
\newcite{LingDyer15ovwr} to LM and POS tagging,
by \newcite{lampe2016} to NER,
by \newcite{BallesterosDyer15chlstm} 
to parsing
morphologically rich
languages, 
and by \newcite{cao2016} to learning word embeddings.
\newcite{subword16} learn word embeddings
 by representing
words with the average of their character ngrams (subwords)
embeddings. 
Similarly, \newcite{chen2015} extends  \texttt{word2vec} for Chinese  
with joint modeling with characters.

\textbf{Fine-grained entity typing}.
Our task is to infer fine-grained types of KB entities. 
KB completion is an application of this task.
\newcite{yyhs15fig}'s FIGMENT system addresses this
task with only contextual information; they do not use character-level and word-level features of entity names.
\newcite{neelakantan2015inferring}
and \newcite{xie16dkrl} also address a similar task,
but they rely on entity descriptions in KBs, which 
in many settings
are not
available.
The problem of Fine-grained mention typing (FGMT) 
\cite{spaniol2012hyena,ling2012fine,yogatama2015acl,delcorro15finet,attentiveTyper16,partialLabel16} 
is related to our task.
FGMT classifies single \emph{mentions} of named entities
to their context dependent types whereas we attempt to identify all types of a KB \emph{entity}
from the aggregation of all its mentions.
FGMT can still be evaluated in our task by aggregating 
the mention level decisions but as we will show in 
our experiments for one system, i.e., FIGER \cite{ling2012fine}, 
our entity embedding based models are better in entity typing.

\begin{figure}[t!] 
\centering{
\includegraphics[scale=0.50]{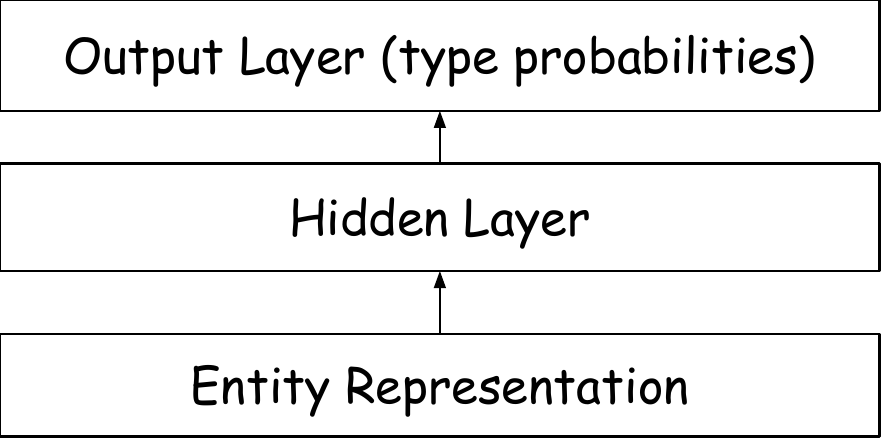}}
\caption{Schematic diagram of our architecture for entity classification.
``Entity Representation'' ($\vec{v}(e)$) is the (one-level or multi-level)
  vector representation of entity. Size of output layer is $|T|$.}
\label{fig:mlp}
\end{figure}

\section{Fine-grained entity typing}
Given (i)
a KB with a set of entities $E$, (ii) a set of types $T$,
and (iii) a large corpus $C$ in which mentions of 
$E$ are
linked, we 
address the 
task of \emph{fine-grained entity typing} \cite{yyhs15fig}: 
predict whether  
entity $e$ is a member of type $t$ or not.
To do so, we use a set of training examples
to learn $P(t|e)$:
the probability that entity $e$ has type $t$.
These probabilities  can be used to assign \emph{new types} to
entities covered in the KB as well as typing \emph{unknown entities}.

We learn $P(t|e)$ with a general architecture; see  \figref{mlp}.
The output layer has size $|T|$.
Unit $t$ of this layer outputs 
the probability for type $t$. 
``Entity Representation'' ($\vec{v}(e)$) is the vector representation of entity $e$ -- we
will describe in detail in the rest of this section what
forms 
$\vec{v}(e)$ takes.
We model $P(t|e)$ as a multi-label classification, and 
train
a multilayer perceptron (MLP) with one hidden layer:
\begin{equation} 
\eqlabel{MLPscore}
\big[ P(t_1|e) \ldots P(t_T|e) \big]
= \sigma\Big(\textbf{W}\dnrm{out}  f\big(\textbf{W}\dnrm{in}\vec{v}(e)\big)\Big)
\end{equation}
where 
$\textbf{W}\dnrm{in} \in \mathbb{R}^{h\times d} $ is the weight matrix from
$\vec{v}(e) \in \mathbb{R}^d$ to the hidden layer with size $h$. 
$f$ is the rectifier function. 
$\textbf{W}\dnrm{out} \in \mathbb{R}^{|T| \times h} $ is the weight matrix
from hidden layer to output layer of size $|T|$.
$\sigma$ is the sigmoid function.
Our objective is binary cross entropy summed over types:
\begin{equation*}
\sum_{t}{-\Big(m_t \log{p_t} + 
(1 - m_t) \log{(1 - p_t)} \Big)}
\end{equation*}
where $m_t$ is the truth and
$p_t$ the prediction.

The key difficulty when trying to  compute $P(t|e)$
is in 
learning a good representation for entity $e$. 
We make use of contexts and name
of $e$
to represent its feature vector
on the three levels of entity, word and character.

\subsection{Entity-level  representation}
\label{ssec:context}
Distributional representations or embeddings are commonly used
for words. 
The underlying hypothesis is that words with similar
meanings tend to occur in similar contexts \cite{harris54}
and therefore cooccur with similar context words.
We can extend the distributional hypothesis to entities (cf.\ \newcite{Wang14joint}, \newcite{yyhs15fig}):
entities with similar meanings tend to have similar contexts.
Thus, we can learn a $d$ dimensional embedding $\vec{v}(e)$ of entity 
$e$ from a corpus in which all mentions of the entity have been replaced by
a special identifier.
We refer to 
these entity vectors as the \emph{entity level representation} (ELR).

In previous work, order information of context words (relative position 
of words in the contexts)
 was generally ignored and objectives similar to the
SkipGram (henceforth: \emph{SKIP}) model were
used to learn $\vec{v}(e)$.  However, 
the bag-of-word context is difficult to distinguish for
pairs of types like (restaurant,food) and
(author,book). This suggests that \emph{using order aware
embedding models is important for entities}.
Therefore, we apply \newcite{wang2vec15}'s extended
version of SKIP, Structured SKIP (SSKIP). It
incorporates the order of context
words into the objective.  We compare it with SKIP
embeddings in our experiments.

\subsection{Word-level  representation}
\label{sec:wlr}
Words inside entity names are important sources
of information for typing entities. 
We define the word-level representation (WLR)
as the \emph{average 
of the embeddings of the words} that the entity name contains $
\vec{v}(e) = 1/n \sum_{i=1}^n \vec{v}(w_i)
$
\noindent where
$\vec{v}(w_i)$ is the embedding
of 
the $i\uprm{th}$ word of an entity name of length
$n$. 
We opt for simple averaging since entity names
often consist of a small number of words with
clear semantics. Thus, averaging is a promising way of
combining the information that each word contributes.

The word embedding, $\vec{w}$,
itself can be learned from models with different granularity levels.
Embedding models that consider words as atomic units in
the corpus, e.g., SKIP and SSKIP,
are word-level.
On the other hand, embedding models 
that represent words with their character ngrams, 
e.g., \texttt{fasttext}  \cite{subword16},
are subword-level.
Based on this, we consider and evaluate \textbf{word-level WLR (WWLR)}
and \textbf{subword-level WLR (SWLR)} in this
paper.\footnote{Subword models have properties of
both  character-level models (subwords are character ngrams) and
of word-level models (they do not cross boundaries between
words). They probably could be put in either category, but
in our context fit the word-level category better
because we see the granularity level with respect to the entities 
and not words.}

\begin{figure}[htp] 
\centering{
\includegraphics[scale=0.40]{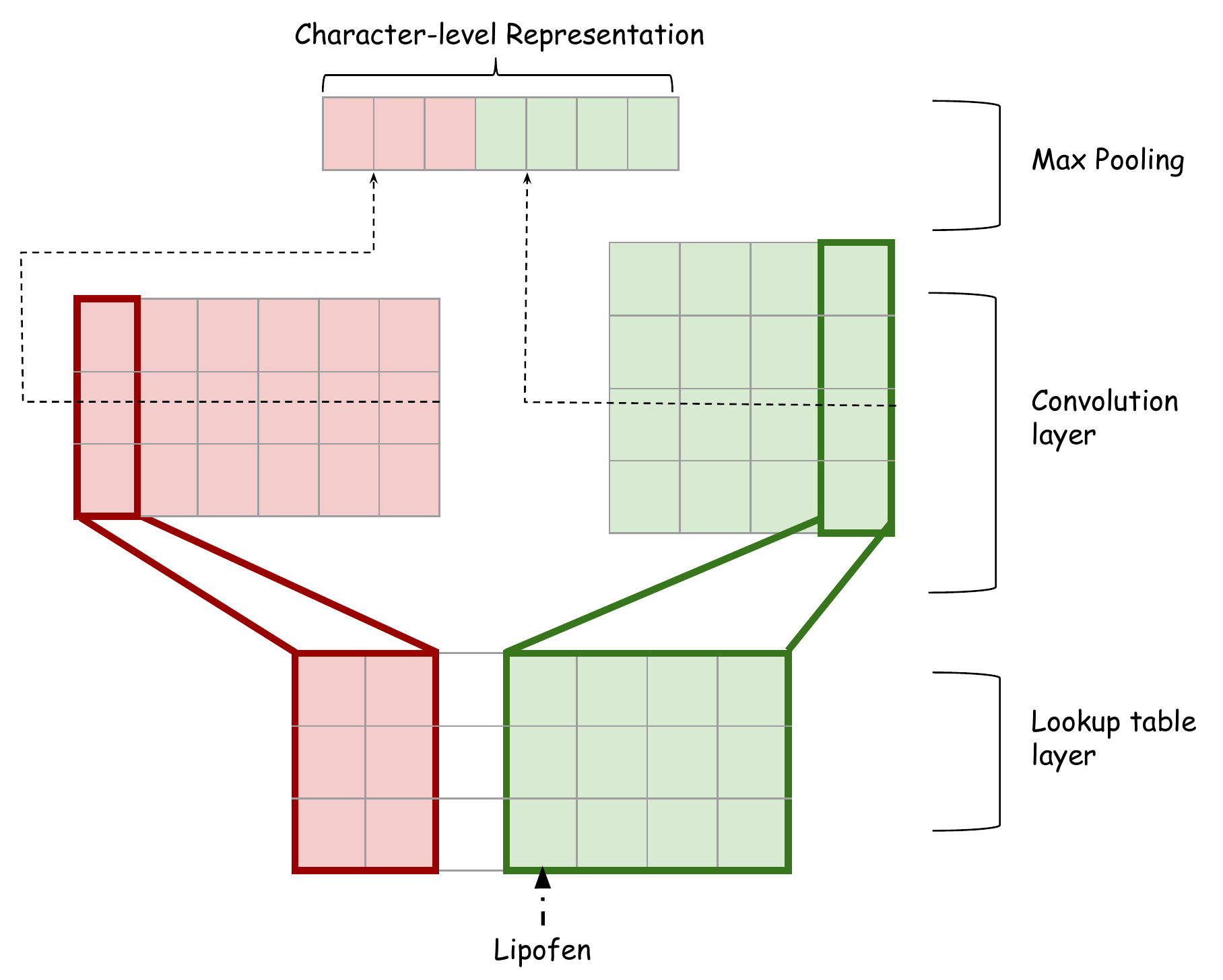}}
\caption{Example architecture for the 
character-level CNN with max pooling.
The input is ``Lipofen''.
Character embedding size is three.
There are three filters of width 2 and 
four filters of width  4.}
\label{fig:cnn}
\end{figure}

\subsection{Character-level representation}
\label{sec:name2type} 
For computing the \emph{character level representation} (CLR), 
we design models that try to type an entity based on the sequence of
characters of its name.  Our hypothesis is that names of
entities of a specific type often have similar character
patterns.  
Entities of type \textsc{ethnicity} often
end in ``ish'' and ``ian'', e.g., ``Spanish'' and
``Russian''. Entities of type \textsc{medicine} often end in
``en'': ``Lipofen'', ``acetaminophen''. 
Also, some types tend to have specific cross-word shapes in
their entities, e.g., 
\textsc{person} names usually consist of two words, 
or \textsc{music} names are usually long, containing several words.

The first layer of the character-level models is a
\emph{lookup table} that maps each character to an
embedding of size $d_c$. These embeddings capture
similarities between characters, e.g., similarity in type of
phoneme encoded (consonant/vowel) or similarity in case
(lower/upper).
The output of the lookup layer for an entity name
is a matrix $C \in \mathbb{R}^{l
  \times d_c}$ where $l$ is the maximum length of a name and
all names are padded to length $l$. This length $l$ includes
special start/end characters that bracket the entity name.

We experiment with four architectures to produce
character-level representations in this paper: FORWARD
(direct forwarding of character embeddings), CNNs,
LSTMs and BiLSTMs. 
The output of each architecture then takes the place
of the entity representation $\vec{v}(e)$ in \figref{mlp}.

\textbf{FORWARD}
simply concatenates all rows
of matrix $C$; thus, $\vec{v}(e) \in \mathbb{R}^{d_c*l}$.

The \textbf{CNN}
uses $k$ filters of different window widths $w$
to narrowly convolve $C$.
For each filter $H \in \mathbb{R}^{d_c\times w}$,
the result of the convolution of $H$ over matrix $C$
is feature map $f \in \mathbb{R}^{l-w+1}$:

$f[i] = \mbox{rectifier}(C_{[:, i : i + w - 1]} \odot H + b)$

\noindent where 
rectifier is the activation function,
$b$ is the bias,
$C_{[:, i : i + w - 1]}$ are the columns $i$ to $i + w - 1$
of $C$,
 $ 1\leq w\leq10$ are the window widths we consider
 and $\odot$   is the sum of element-wise multiplication. 
Max pooling then gives us one feature for each
filter. The concatenation of all these features 
is our representation: 
$\vec{v}(e) \in \mathbb{R}^{k}$.
An example CNN architecture is show in \figref{cnn}.

The input to the \textbf{LSTM}
is the character sequence in matrix $C$, i.e., 
$x_1,\dots ,x_l \in \mathbb{R}^{d_c}$.
It generates the state sequence $h_1, . . . ,h_{l+1}$ and 
the output is the last state $\vec{v}(e) \in \mathbb{R}^{d_h}$.\footnote{We use Blocks 
\cite{Merrien15blocks}.}

The \textbf{BiLSTM} consists of
two LSTMs, one going forward, one going backward.
The first state of the backward LSTM is initialized as
$h_{l+1}$,
the 
last state of the forward LSTM.
The BiLSTM entity representation is
the concatenation of  last states of forward and backward
LSTMs, i.e.,
$\vec{v}(e) \in \mathbb{R}^{2 * d_h}$.

\begin{figure}[h] 
\centering{
\includegraphics[width=0.45\textwidth,
height=65pt]{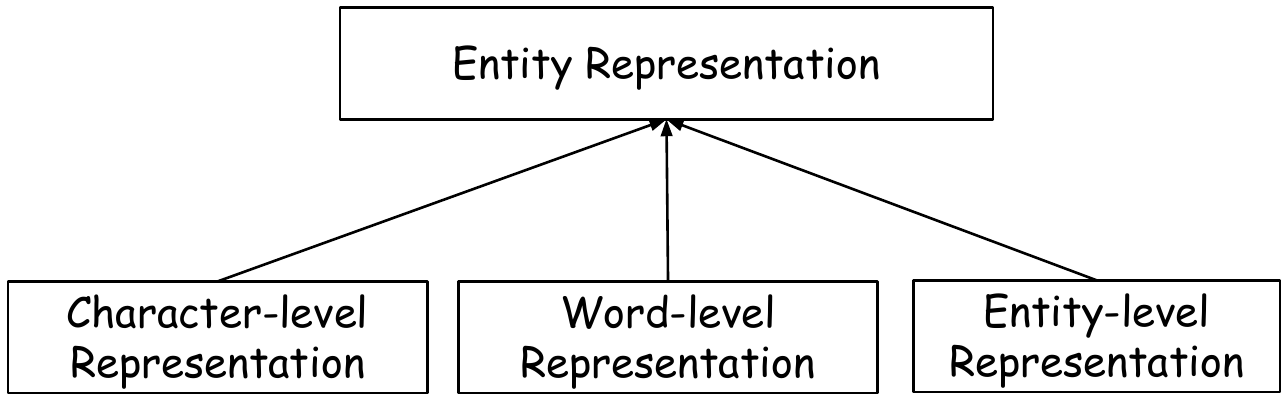}}
\caption{Multi-level representation
}
\label{fig:joint}
\end{figure}

\subsection{Multi-level representations}

Our different levels of representations can
give complementary information 
about entities.

\enote{hs}{i think at some point you need to introduce
  ``cross-word'' and say a sentence about the distintino you
  make bewteen ``subword'' and ``cross-word''}
\textbf{WLR and CLR}. 
Both WLR models, SWLR and WWLR, do not have access to
the cross-word character ngrams 
of entity names while CLR models do.
Also, CLR is task specific by training
on the entity typing dataset
while WLR is generic.
On the other hand, WWLR and SWLR models 
have access to information that CLR ignores:
the tokenization of entity names into words and 
embeddings of these words. 
It is clear that words 
are
particularly important character sequences since they
often correspond to linguistic units with clearly
identifiable semantics -- which is not true for most
character sequences.
For many entities, 
the words they contain are a better basis for typing
than the 
character sequence. For example, even
if ``nectarine'' and ``compote'' did not occur in any names
in the training corpus, we can still learn good word
embeddings from their non-entity occurrences. This then
allows us to correctly type the entity ``Aunt Mary's
Nectarine Compote''
as \textsc{food} based on
the sum of the word embeddings.

\textbf{WLR/CLR and ELR}. 
Representations from entity names, i.e., WLR and CLR, by themselves are limited because 
many classes of names can be used for different types of entities;
e.g., person names do not contain hints as to whether
they are referring to a politician or athlete.
In contrast, the ELR
embedding is based on an
entity's contexts, which are often informative  
 for each entity
and can distinguish politicians from  athletes.
On the other hand, 
not all  entities
have sufficiently many informative contexts in the corpus.
For these entities, their name can be a complementary 
source of information and character/word level 
representations can increase typing accuracy.

Thus, we introduce joint models that use combinations of the
three levels.  Each multi-level model concatenates several
levels.  We train the constituent embeddings as follows.
WLR and ELR are computed as described above and are not
changed during training.  CLR -- produced by one of the
character-level networks described above -- is initialized
randomly and then tuned during training.  Thus, it can focus
on complementary information related to the task that is not
already present in other levels.
The schematic diagram of our multi-level representation is shown 
in \figref{joint}.

\section{Experimental setup and results}
\label{sect:exp}
\subsection{Setup}
\textbf{Entity datasets and corpus.} 
We address the task of fine-grained entity typing and 
use \newcite{yyhs15fig}'s FIGMENT dataset\footnote{\url{cistern.cis.lmu.de/figment/} \label{fc}}
for
evaluation. The FIGMENT 
corpus is part of a version of ClueWeb
in which Freebase entities are annotated using
FACC1 \cite{lemur16url,gabrilovich2013facc1}.
The FIGMENT entity datasets
 contain 200,000 Freebase entities
that were mapped to 102 FIGER types \cite{ling2012fine}.
We use the same train (50\%), dev (20\%) and test (30\%) partitions  as
\newcite{yyhs15fig} and
extract the names from mentions of dataset entities 
in the corpus.
We take the most frequent name for dev and test entities
and three most frequent names for train 
(each one tagged with entity types).

\textbf{Adding parent types to refine entity dataset.} 
FIGMENT ignores that FIGER is a proper hierarchy of types;
e.g.,
while \textsc{hospital}  is a subtype of \textsc{building} according to 
FIGER,  there are entities in FIGMENT that are
hospitals, but not buildings.\footnote{See \url{github.com/xiaoling/figer} for FIGER}
Therefore, we modified the FIGMENT dataset by adding 
for each assigned type (e.g., \textsc{hospital})
its
parents (e.g., \textsc{building}).
This makes FIGMENT more consistent and
eliminates spurious false negatives 
(\textsc{building} in the example).

We now describe our \textbf{baselines}: (i) BOW \& NSL: 
hand-crafted features, (ii) FIGMENT 
\cite{yyhs15fig} and
(iii) adapted version of FIGER \cite{ling2012fine}.

We implement the following two feature sets from the
literature as a \emph{hand-crafted baseline} for
our character and word level 
models.  (i) \emph{BOW}: individual words of entity name (both
as-is and lowercased); (ii) \emph{NSL} (ngram-shape-length): shape and length of the
entity name (cf.\ \newcite{ling2012fine}), character
$n$-grams, $1 \leq n \leq n\dnrm{max}, n\dnrm{max}=5$ (we
also tried $n\dnrm{max}=7$, but results were worse on dev) and
normalized character $n$-grams: lowercased, digits
replaced by ``7'', punctuation replaced by ``.''.
These features are represented as a sparse binary vector 
$\vec{v}(e)$ that is input to the architecture in
\figref{mlp}.

\emph{FIGMENT} is the model for entity typing presented by
\newcite{yyhs15fig}. 
The authors only use entity-level
representations for entities trained by SkipGram, so the FIGMENT baseline
corresponds to the
entity-level result shown as ELR(SKIP) in the tables. 

The third baseline is using an existing mention-level entity typing
system, \emph{FIGER} \cite{ling2012fine}.
FIGER uses a wide variety of  features
on different levels (including
parsing-based features)
from contexts of entity mentions
as well as the mentions themselves and returns a score for each
mention-type instance in the corpus.
We provide the ClueWeb/FACC1 segmentation of entities,
so FIGER does not need to recognize entities.\footnote{Mention typing is separated from recognition in FIGER model. So it can use our segmentation of entities.}
We use the trained model provided by the authors
and normalize FIGER scores using
softmax to make them comparable for aggregation.
We experimented with different aggregation functions
(including maximum and k-largest-scores for a type), but
 we use the average of scores since it gave us the best result on dev.
 We call this baseline AGG-FIGER.

\textbf{Distributional embeddings.}
For WWLR and ELR, 
we use SkipGram model in \texttt{word2vec} and SSkip model in \texttt{wang2vec}
\cite{wang2vec15} to learn 
embeddings for words, entities and types.
To obtain embeddings for all three
in the same space, we process ClueWeb/FACC1  as follows.
For each sentence $s$,
we add three copies: $s$ itself, a copy of $s$ in which each
entity is replaced with its Freebase identifier (MID) and a
copy in which each entity (not test entities though) is replaced with an ID indicating
its notable type.
The resulting corpus contains around
4 billion tokens and 1.5 billion types.

We run  
SKIP and SSkip  with the same setup
(200 dimensions, 10 negative samples, window size 5, word frequency threshold of 100)\footnote{The threshold does not apply for MIDs.}
on this corpus to learn embeddings for words,
entities and FIGER types.
Having entities and types in the same vector space,
we can add another feature vector $\vec{v}(e) \in
\mathbb{R}^{|T|}$ (referred to as TC below):
for each entity, we compute cosine similarity of its entity vector 
with all type vectors. 

For SWLR, 
we use \texttt{fasttext}\footnote{\url{github.com/facebookresearch/fastText}}
to learn word embeddings from the ClueWeb/FACC1 corpus.
We use similar settings as our WWLR SKIP and SSkip embeddings and
keep the defaults of other hyperparameters.
Since the trained model of \texttt{fasttext} is applicable for 
new words, we apply the model to get 
embeddings for the filtered rare words as well. 

\begin{table}[tbhp]
\begin{center}
{
\scriptsize
\begin{tabular}{lc}
model  & hyperparameters\\ 
\hline
CLR(FF)      & $d_{c}=15, h_{mlp}=600$   \\
\hline
CLR(LSTM)      & $d_{c}=70, d_{h}=70, h_{mlp}=300$ \\
\hline
CLR(BiLSTM)   & $d_{c}=50, d_{h}=50, h_{mlp}=200$ \\
\hline
CLR(CNN)    & $d_{c}=10, w=[1,..,8]$ \\
		   & $n=100, h_{mlp}=800$  \\   

\hline
CLR(NSL)    & $h_{mlp}=800$  \\   

\hline
BOW    & $h_{mlp}=200$  \\   
\hline
BOW+CLR(NSL) 
            & $h_{mlp}=300$  \\   

\hline
WWLR        & $h_{mlp}=400$\\
\hline
SWLR        & $h_{mlp}=400$\\

\hline
WWLR+CLR(CNN)  
		   & $w=[1,...,7]$ \\ 
           & $d_{c}=10, n=50, h_{mlp}=700$ \\
\hline
SWLR+CLR(CNN)  
		   & $w=[1,...,7]$ \\ 
           & $d_{c}=10, n=50, h_{mlp}=700$ \\
\hline
ELR(SKIP)        & $h_{mlp}=400$\\
\hline
ELR(SSKIP)        & $h_{mlp}=400$\\

\hline
ELR+CLR   & $d_{c}=10, w=[1,...,7]$\\
&                $n=100, h_{mlp}=700$\\

\hline
ELR+WWLR    & $h_{mlp}=600$\\
\hline
ELR+SWLR    & $h_{mlp}=600$\\
\hline
ELR+WWLR+CLR 
			& $d_{c}=10, w=[1,...,7]$\\
&                $n=50, h_{mlp}=700$\\
\hline
ELR+SWLR+CLR 
			& $d_{c}=10, w=[1,...,7]$\\
&                $n=50, h_{mlp}=700$\\
\hline
ELR+WWLR+CNN+TC
                 & $d_{c}=10, w=[1,...,7]$\\
&                $n=50, h_{mlp}=900$ \\
\hline
ELR+SWLR+CNN+TC(MuLR)
                 & $d_{c}=10, w=[1,...,7]$\\
&                $n=50, h_{mlp}=900$ \\
\hline
AVG-DES
                & $h_{mlp}=400$ \\
\hline
MuLR+AVG-DES
                 & $d_{c}=10, w=[1,...,7]$\\
&                $n=50, h_{mlp}=1000$ \\
\hline

\end{tabular} 
}
\caption{Hyperparameters of different models.
$w$ is the filter size. 
$n$ is the number of CNN feature maps for each filter size. 
$d_c$ is the character embedding size.
$d_h$ is the LSTM hidden state size. 
$h\dnrm{mlp}$ is the number of hidden units in the MLP.
}
\tablabel{hyp}
\end{center}
\end{table}

Our \textbf{hyperparameter
values} are given in  \tabref{hyp}.
The values are optimized on dev.
We use AdaGrad and minibatch training.
For each experiment, we select the best model on dev.

We use 
these \textbf{evaluation measures}:
(i) accuracy: an entity is correct if
all its types and no incorrect types are assigned to it;
(ii) micro average $F_1$: $F_1$ of all type-entity assignment decisions;
(iii) entity macro average $F_1$: $F_1$ of types assigned to
an entity, averaged over entities; 
(iv) type macro average $F_1$: $F_1$ of entities assigned to
a type, averaged over types.

The assignment decision 
is  based on thresholding the probability function
$P(t|e)$. For each model and type,
we select
the threshold that maximizes 
$F_1$ of entities assigned
to the type on dev.

\def \hfillx {\hspace*{-\textwidth} \hfill}

\def\makesmaller{0.08cm}

\begin{table*}[t]
\begin{center}
\setlength{\tabcolsep}{3pt}
{
{\footnotesize
\begin{minipage}{0.55\textwidth}

\begin{tabular}{rl|ccc|ccc|ccc}
&& \multicolumn{3}{|c|}{all entities} &
  \multicolumn{3}{|c|}{head entities} &
  \multicolumn{3}{|c}{tail entities}\\
&& acc & mic &  mac 
& acc &  mic &  mac 
& acc &  mic &  mac\\ 
\hline\hline %
1&MFT    
         & .000 & .041 & .041
	     & .000 & .044 & .044   
    	 & .000 & .038 & .038 \\
\hline
      
2&CLR(FORWARD)
         & .066 & .379 & .352 
         & .067 & .342 & .369 
         & .061 & .374 & .350\\ 
         
3&CLR(LSTM)   
		 & .121 & .425 & .396 
         & .122 & .433 & .390 
         & .116 & .408 & .391\\ 

4&CLR(BiLSTM)  
		 & .133 & .440 & .404 
         & .129 & .443 & .394 
		 & .135 & .428 & .404 \\

5&CLR(NSL)
		 & .164 & .484 & .464
         & .157 & .470 & .443
         & .173 & .483 & .472 \\ 
6&CLR(CNN)   
         & .177 & .494 & .468
         & .171 & .484 & .450 
         & .187 & .489 & .474\\ 

\hdashline
7&BOW
		 & .113 & .346 & .379
         & .109 & .323 & .353 
         & .120 & .356 & .396\\
8&WWLR(SKIP)
		 & .214 & .581 & .531
         & .293 & .660 & .634 
         & .173 & .528 & .478\\ 
9&WWLR(SSKIP)
		 & .223 & .584 & .543
         & .306 & .667 & .642 
         & .183 & .533 & .494\\ 
10&SWLR
		 & .236 & .590 & .554
         & .301 & .665 & .632 
         & .209 & .551 & .522\\

\hdashline         

11&BOW+CLR(NSL)
		 & .156 & .487 & .464
         & .157 & .480 & .452 
         & .159 & .485 & .469\\
12&WWLR+CLR(CNN)
		 & .257 & .603 & .568
         & .317 & .668 & .637 
         & .235 & .567 & .538\\

13&SWLR+CLR(CNN)
		 & .241 & .594 & .561
         & .295 & .659 & .628 
         & .227 & .560 & .536\\
         
\hdashline
14&ELR(SKIP)   
		 & .488 & .774 & .741 
         & .551 & .834 & .815 
         & .337 & .621 & .560\\

15&ELR(SSKIP)  
		 & .515 & .796 & .763 
         & .560 & .839 & .819 
         & .394 & .677 & .619\\

\hdashline
16&AGG-FIGER  
        & .320 & .694 & .660 
		& .396 & .762 & .724 
        & .220 & .593 & .568\\ 

17&ELR+CLR  
        & .554 & .816 & .788 
		& .580 & .844 & .825
        & .467 & .733 & .690\\ 
18&ELR+WWLR
           & .557 & .819 & .793
           & .582 & .846 & .827
           & .480 & .749 & .708\\

19&ELR+SWLR
           & .558 & .820 & .796
           & .584 & .846 & .829
           & .480 & .751 & .714\\
           
20&ELR+WWLR+CLR
		   & .568 & .823 & .798
           & .590 & .847 & .829
           & .491 & .755 & .716\\
21&ELR+SWLR+CLR
		   & .569 & .824 & .801
           & .590 & .849 & \textbf{.831}
           & .497 & .760 & .724\\
22&ELR+WWLR+CLR+TC
		   & .572 & .824 & .801
           & .594 & .849 & \textbf{.831}
           & .499 & .759 & .722\\
               
23&ELR+SWLR+CLR+TC
		   & \textbf{.575} & \textbf{.826} & \textbf{.802}
           & \textbf{.597} & \textbf{.851} &\textbf{.831}
           & \textbf{.508} & \textbf{.762} & \textbf{.727}\\

%

\end{tabular} 
 
\caption{
Accuracy (acc),  micro (mic) and macro (mac) $F_1$ on test for all,
head and tail entities.}

\tablabel{big}
\end{minipage}
\hfill
\begin{minipage}{0.3\textwidth}

\vspace{-1cm}

\begin{tabular}{l|c|c|c}
        \multicolumn{1}{r|}{types:}      & all     & head     & tail \\ 
\hline
AGG-FIGER         & .566        & .702          & .438\\
ELR               & .621         & .784          & .480\\ 
MuLR   &\textbf{.669} & \textbf{.811} & \textbf{.541}\\
\end{tabular} 
\caption{Type macro average $F_1$ on test for all, head and
  tail types.
MuLR = ELR+SWLR+CLR+TC}
\tablabel{typemacro}
\smallskip
\begin{tabular}{l|l@{\hspace{\makesmaller}}l@{\hspace{\makesmaller}}l}
     &   all & \multicolumn{2}{c}{known?} \\ 
  	   &  & \multicolumn{1}{c}{yes} & \multicolumn{1}{c}{no} \\ 
\hline
CLR(NSL)        & .484 & .521  & .341\\ 
CLR(CNN)        & .494 & .524  & .374\\ 
\hdashline
BOW             & .346 & .435  & .065 \\
SWLR            & .590 & .612  & .499\\
\hdashline
BOW+NSL         & .497 & .535  & .358 \\

SWLR+CLR(CNN)
          & \textbf{.594} & \textbf{.616}   & \textbf{.508}

\end{tabular}
\caption{Micro $F_1$ on test of character, word level models for
  all, known (``known? yes'') and unknown (``known? no'') entities.
}
\tablabel{unknown}
\end{minipage}
}
}
\end{center}
\end{table*}

\subsection{Results}
\tabref{big} gives results on the test entities
for all  (about 60,000 entities), 
head (frequency $>$
100; about 12,200) and tail (frequency $<$ 5; about 10,000).
\emph{MFT} (line 1) is the most frequent type baseline that
ranks types according to their frequency in
the train entities.
Each level of representation is separated with dashed lines,
and -- unless noted otherwise -- the best of each level is joined in multi level
representations.\footnote{For accuracy measure: in the following ordered lists
  of sets, $A$$<$$B$ means that all members (row numbers in \tabref{big}) of $A$ are
  significantly worse than all members of $B$:
\{1\} $<$ \{2\} $<$ \{3, \dots, 11\} $<$ \{12,13\} $<$ \{14,15,16\} 
$<$\{17, \dots, 23\}.
Test of equal proportions, $\alpha<$ 0.05.
See \tabref{sigTable} in the appendix
for more details.
}

\todo{
Accuracies that are not significant (p.value $>$ 0.05):
format: MODEL1 * MODEL2 all/head/tail
 PLEASE ADD NOSIG TO GIT
\input{nosig.acc.txt} 
}

\textbf{Character-level models} 
are on lines 2-6. 
The order of systems is:
CNN $>$ NSL  $>$ BiLSTM $>$ LSTM $>$ FORWARD. 
The results show that  complex 
neural networks are more effective than simple forwarding.
BiLSTM works better than LSTM, confirming 
other related work.
CNNs probably work better than LSTMs
because there are
few complex 
non-local dependencies in the sequence, but
many important local features.
CNNs with maxpooling can more straightforwardly
capture local and position-independent features.
CNN also beats NSL baseline; a possible reason is
that
CNN -- an automatic method of feature learning -- is more robust
than hand engineered feature based NSL.
We show  more detailed results in 
\secref{analysis}.

\textbf{Word-level models} are on lines 7-10.  
BOW performs
worse than WWLR because it cannot deal well with sparseness.
SSKIP uses word order information in WWLR and performs better than SKIP.
SWLR uses subword information and
performs better than WWLR,
especially for tail entities.
Integrating subword information improves
the quality of embeddings for rare words
and mitigates the problem of unknown words.

\textbf{Joint word-character level models}
are on lines 11-13.
WWLR+CLR(CNN) and SWLR+CLR(CNN) beat the component models.
This confirms our underlying assumption in designing
the complementary multi-level models.
BOW problem with rare words does not allow its joint model with NSL
to work better than NSL.
WWLR+CLR(CNN) works better than BOW+CLR(NSL) by 10\% micro $F_1$,
again due to the limits of BOW compared to WWLR.
Interestingly WWLR+CLR works better than SWLR+CLR
and this suggests that WWLR is indeed richer than SWLR
when
CLR  mitigates its problem  with rare/unknown words 

\begin{figure}[ht] 
\centering{
\includegraphics[width=200pt,height=180pt]{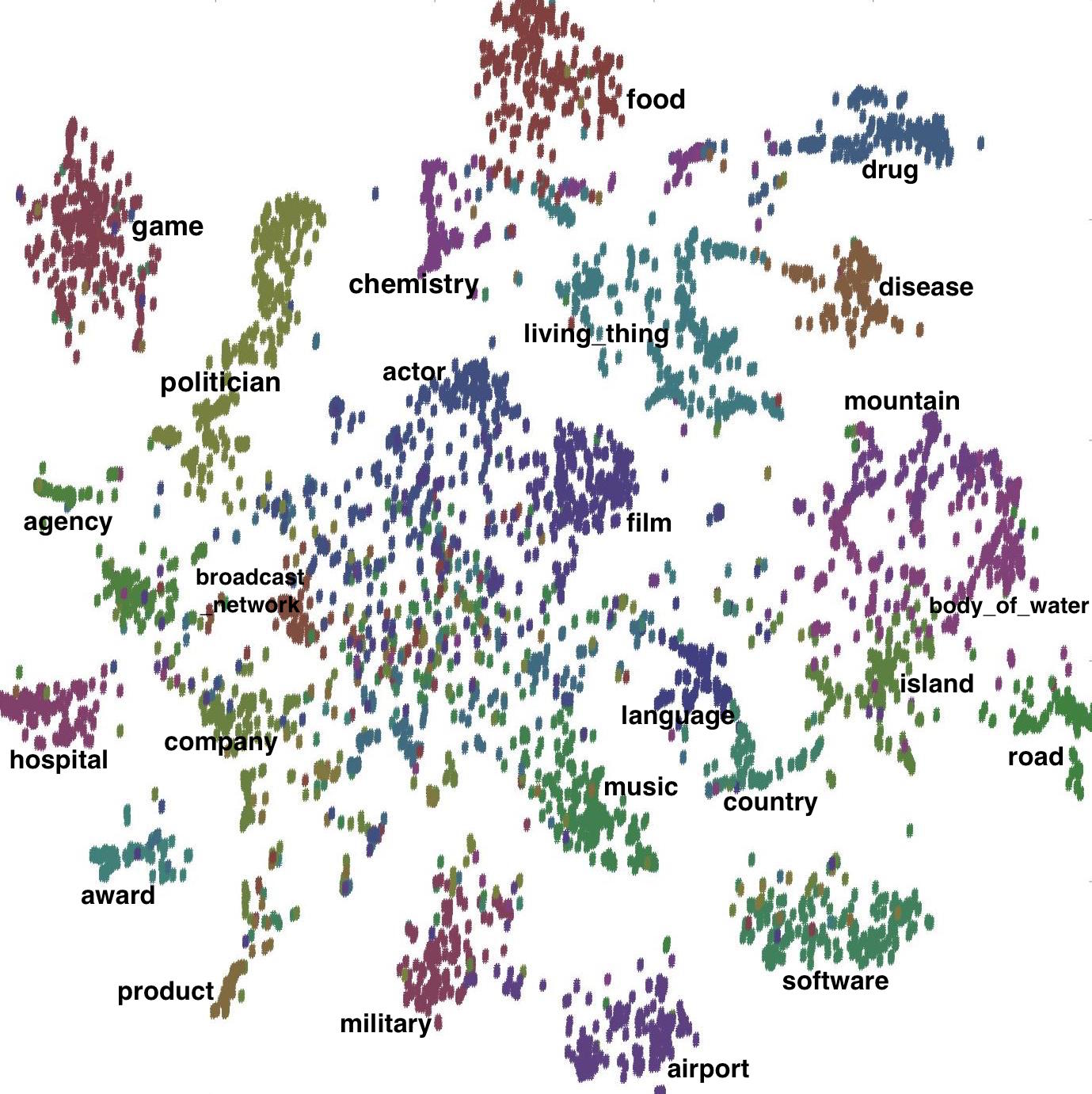}}
\caption{t-SNE result of entity-level representations} 
\label{fig:tsne}
\end{figure}

\textbf{Entity-level models} are on lines 14--15
and they are 
better than all previous models on lines 1--13.
This shows the power of 
entity-level embeddings.
In \figref{tsne},
a t-SNE \cite{van08tsne} visualization of ELR(SKIP)
embeddings
using different colors for entity types
shows that entities of the same type are clustered together.
SSKIP works marginally better than SKIP for ELR, especially for tail entities, 
confirming our hypothesis that order information is important for 
a good distributional entity representation.
This is also confirming the results of \newcite{derata16acl},
where they  also get better entity typing results with SSKIP compared to SKIP. They propose to use entity typing as an extrinsic evaluation 
for embedding models.  

\textbf{Joint entity, word, and
character level models}
are on lines 16-23. 
The AGG-FIGER baseline works better than the systems on lines 1-13,
but worse than ELRs.
This is probably due to the
fact that AGG-FIGER is optimized for mention typing
and it is trained using distant supervision assumption. 
Parallel to our work, \newcite{ourjoint2017} optimize a mention typing
model for our entity typing task by introducing multi instance learning
algorithms,
resulting  comparable performance to ELR(SKIP).
We will investigate their method in future.

Joining CLR
with ELR (line 17) results in
large improvements, especially for tail entities (5\% micro
$F_1$).  
This demonstrates that for rare entities,
contextual information is often not sufficient for an
informative representation, hence name features are important.
This is also true for the joint models of WWLR/SWLR and ELR (lines 18-19).
Joining WWLR works better than CLR, 
and SWLR is slightly better than WWLR.
Joint models of WWLR/SWLR with ELR+CLR gives more improvements,
and SWLR is again slightly better than WWLR.
ELR+WWLR+CLR and ELR+SWLR+CLR,
are better than their two-level counterparts, 
again confirming that these levels are complementary.

We get a further boost, especially for tail entities, 
by also including TC (type
cosine) in the combinations (lines 22-23).
This demonstrates the
potential advantage of having a common representation space
for entities and types.
Our best model, ELR+SWLR+CLR+TC (line 22), which we refer to as MuLR
in the other tables,
beats our initial baselines (ELR and AGG-FIGER) by large margins, 
e.g., in tail entities  
improvements are more than 8\% in micro F1.

\tabref{typemacro} shows \textbf{type macro $F_1$ }   for 
MuLR (ELR+SWLR+CLR+TC) and
two baselines.
There are 11 head types (those with $\geq$3000 train
entities) and 36 tail types (those with $<$200 
train entities). 
These results again confirm the superiority of
our multi-level models over the baselines: AGG-FIGER
and ELR, the best single-level model baseline.

\subsection{Analysis}
\seclabel{analysis}

\textbf{Unknown vs.\ known entities.}
To analyze the complementarity of character and
word level representations,
as well as more fine-grained comparison of our
models and the baselines,
we divide  test entities into 
\emph{known entities} -- at least one word of the entity's name
appears in a train entity -- and \emph{unknown entities} (the complement).
There are 45,000 (resp.\ 15,000) known (resp.\ unknown) test entities.

\tabref{unknown}  shows that the
CNN works only slightly better (by 0.3\%) than NSL on  known entities,
but works
much better 
on unknown entities (by 
3.3\%), justifying our preference for deep learning CLR models.
As expected, BOW works 
relatively well for known entities and
really poorly for unknown entities.
SWLR beats CLR models as well as BOW.
The reason is that in our setup, 
word embeddings are induced on the entire corpus using an unsupervised 
algorithm.
Thus, even for many words that
did not occur in
train, SWLR has access to informative representations of words.
The joint model, SWLR+CLR(CNN), is significantly better
than BOW+CLR(NSL) again due to limits of BOW.
SWLR+CLR(CNN) is better than SWLR in unknown entities.

\textbf{Case study of \textsc{living-thing}}.
To understand the interplay of different levels better, we
perform a 
case study of the type
\textsc{living-thing}. Living beings that are not humans
belong to this type.

WLRs incorrectly 
assign 
``Walter Leaf'' (\textsc{person}) and ``Along Came A Spider'' (\textsc{music})
to
\textsc{living-thing} because these
names contain a word  referring to a 
\textsc{living-thing} (``leaf'', ``spider''), but
the entity itself is not a 
\textsc{living-thing}. In these cases, the averaging of
embeddings that WLR performs is misleading.
The CLR(CNN) types these two entities  correctly  because 
their names contain character ngram/shape patterns   that are 
indicative of \textsc{person} and \textsc{music}.

ELR incorrectly assigns 
``Zumpango'' (\textsc{city}) and
``Lake Kasumigaura'' (\textsc{location}) to
\textsc{living-thing} because
these entities are rare and words associated with living things (e.g.,
``wildlife'') dominate in their
contexts.
However, CLR(CNN) and WLR enable
the joint model to type the two entites correctly:
``Zumpango'' because of the informative suffix ``-go'' and
``Lake Kasumigaura'' because of the informative word
``Lake''.

While some of the \textbf{remaining errors} of our best system
MuLR are due to the inherent difficulty of
entity typing (e.g., it is difficult to correctly
type a one-word entity that occurs once and whose name is
not informative), many other errors are due to artifacts of
our setup.
First, 
ClueWeb/FACC1 
is the result of an automatic entity linking
system and any entity linking errors propagate to our models.
Second, 
due to the incompleteness of Freebase \cite{yyhs15fig},
many entities in
the FIGMENT dataset are incompletely annotated, resulting in
correctly typed entities being evaluated as incorrect.

\textbf{Adding another source: description-based embeddings.}
While in this paper, we focus on the contexts and names of entities,
there is a textual source of information about entities in KBs
which we can also make use of:
descriptions of entities. 
We extract Wikipedia descriptions of FIGMENT entities 
filtering out the entities ($\sim$ 40,000 out of $\sim$ 200,000) without description.

We then build a simple entity representation 
by averaging the embeddings of the top $k$  words (wrt tf-idf)  of the description (henceforth, AVG-DES).\footnote{$k$ = 20 gives the best results on dev.}
This representation is used as input in \figref{mlp}
to train the MLP.
We also train our best multi-level model as well as the joint of the two  
on this smaller dataset.
Since the descriptions are coming from Wikipedia, we
use 300-dimensional Glove \cite{glove16url} 
embeddings pretrained on Wikipdia+Gigaword 
to get more coverage of words.
For MuLR, we still use the embeddings we trained before.

Results are shown in \tabref{desc}. 
While for head entities, MuLR works marginally better, the difference is
very small in tail entities. 
The joint model of the two (by concatenation of vectors) improves 
the micro F1, with clear boost for tail entities. 
This suggests that for tail entities, the contextual and name information
is not enough by itself and some keywords from descriptions can be really helpful.
Integrating more complex description-based embeddings, e.g., by using CNN \cite{xie16dkrl}, 
may improve the results further.
We leave it for future work.

\begin{table}[t]
\footnotesize{
\begin{tabular}{l|c|c|c}
        \multicolumn{1}{r|}{entities:}      & all     & head     & tail \\ 
\hline
AVG-DES             &.773 & .791 & .745 \\
MuLR                &.825 & .846 & .757 \\
MuLR+AVG-DES  
				    &  \textbf{.873} & \textbf{.877}  & \textbf{.852}\\
\end{tabular} 
\caption{Micro average $F_1$ results of MuLR and description based
model and their joint. 
}
\tablabel{desc}
}
\end{table}

\section{Conclusion}
In this paper, we have introduced representations of
entities on different levels: character, word and entity.
The character level representation is learned from the entity
name. The word level representation is computed from the 
embeddings of the words $w_i$ in the entity name where the
embedding of $w_i$ is derived from
the corpus contexts of $w_i$.  
The entity level representation 
of entity $e_i$
is derived from
the corpus contexts of $e_i$.  
Our
experiments show that each of these levels contributes
complementary information for the task of fine-grained
typing of entities.  The joint model of all three levels
beats the state-of-the-art baseline 
by large margins.
We further showed that extracting some keywords from Wikipedia descriptions of entities, when available,
can considerably improve entity representations, especially
for rare entities.
We believe that our findings can be transferred to other 
tasks where entity representation matters.

\textbf{Acknowledgments.} This work was supported by DFG
(SCHU 2246/8-2).

\bibliography{ref.bib}
\bibliographystyle{eacl2017}

\appendix
\section{Supplementary Material}

\begin{table*}[tbhp]
\begin{center}
\scriptsize
\begin{tabular}{l}
\textbf{All entities}\\
\begin{tabular}{llp{0.01cm}p{0.01cm}p{0.01cm}p{0.01cm}p{0.01cm}p{0.01cm}p{0.01cm}p{0.01cm}p{0.01cm}p{0.01cm}p{0.01cm}p{0.01cm}p{0.01cm}p{0.01cm}p{0.01cm}p{0.01cm}p{0.01cm}p{0.01cm}p{0.01cm}p{0.01cm}p{0.01cm}p{0.01cm}p{0.01cm}}
  &       Models  &01& 02& 03& 04& 05& 06& 07& 08& 09& 10& 11& 12& 13& 14& 15& 16& 17& 18& 19& 20& 21& 22& 23 \\
\hline
01 &           MFT & 0 & 0 & 0 & 0 & 0 & 0 & 0 & 0 & 0 & 0 & 0 & 0 & 0 & 0 & 0 & 0 & 0 & 0 & 0 & 0 & 0 & 0 & 0 \\
02 &  CLR(FORWARD) & * & 0 & 0 & 0 & 0 & 0 & 0 & 0 & 0 & 0 & 0 & 0 & 0 & 0 & 0 & 0 & 0 & 0 & 0 & 0 & 0 & 0 & 0 \\
03 &     CLR(LSTM) & * & * & 0 & 0 & 0 & 0 & * & 0 & 0 & 0 & 0 & 0 & 0 & 0 & 0 & 0 & 0 & 0 & 0 & 0 & 0 & 0 & 0 \\
04 &   CLR(BiLSTM) & * & * & * & 0 & 0 & 0 & * & 0 & 0 & 0 & 0 & 0 & 0 & 0 & 0 & 0 & 0 & 0 & 0 & 0 & 0 & 0 & 0 \\
05 &      CLR(CNN) & * & * & * & * & 0 & * & * & 0 & 0 & 0 & * & 0 & 0 & 0 & 0 & 0 & 0 & 0 & 0 & 0 & 0 & 0 & 0 \\
06 &      CLR(NSL) & * & * & * & * & 0 & 0 & * & 0 & 0 & 0 & * & 0 & 0 & 0 & 0 & 0 & 0 & 0 & 0 & 0 & 0 & 0 & 0 \\
07 &           BOW & * & * & 0 & 0 & 0 & 0 & 0 & 0 & 0 & 0 & 0 & 0 & 0 & 0 & 0 & 0 & 0 & 0 & 0 & 0 & 0 & 0 & 0 \\
08 &   WWLR(SkipG) & * & * & * & * & * & * & * & 0 & 0 & 0 & * & 0 & 0 & 0 & 0 & 0 & 0 & 0 & 0 & 0 & 0 & 0 & 0 \\
09 &  WWLR(SSkipG) & * & * & * & * & * & * & * & * & 0 & 0 & * & 0 & 0 & 0 & 0 & 0 & 0 & 0 & 0 & 0 & 0 & 0 & 0 \\
10 &          SWLR & * & * & * & * & * & * & * & * & * & 0 & * & 0 & 0 & 0 & 0 & 0 & 0 & 0 & 0 & 0 & 0 & 0 & 0 \\
11 &  BOW+CLR(NSL) & * & * & * & * & 0 & 0 & * & 0 & 0 & 0 & 0 & 0 & 0 & 0 & 0 & 0 & 0 & 0 & 0 & 0 & 0 & 0 & 0 \\
12 & WWLR+CLR(CNN) & * & * & * & * & * & * & * & * & * & * & * & 0 & * & 0 & 0 & 0 & 0 & 0 & 0 & 0 & 0 & 0 & 0 \\
13 & SWLR+CLR(CNN) & * & * & * & * & * & * & * & * & * & * & * & 0 & 0 & 0 & 0 & 0 & 0 & 0 & 0 & 0 & 0 & 0 & 0 \\
14 &    ELR(SkipG) & * & * & * & * & * & * & * & * & * & * & * & * & * & 0 & 0 & * & 0 & 0 & 0 & 0 & 0 & 0 & 0 \\
15 &   ELR(SSkipG) & * & * & * & * & * & * & * & * & * & * & * & * & * & * & 0 & * & 0 & 0 & 0 & 0 & 0 & 0 & 0 \\
16 &     AGG-FIGER & * & * & * & * & * & * & * & * & * & * & * & * & * & 0 & 0 & 0 & 0 & 0 & 0 & 0 & 0 & 0 & 0 \\
17 &       ELR+CLR & * & * & * & * & * & * & * & * & * & * & * & * & * & * & * & * & 0 & 0 & 0 & 0 & 0 & 0 & 0 \\
18 &      ELR+WWLR & * & * & * & * & * & * & * & * & * & * & * & * & * & * & * & * & 0 & 0 & 0 & 0 & 0 & 0 & 0 \\
19 &      ELR+SWLR & * & * & * & * & * & * & * & * & * & * & * & * & * & * & * & * & 0 & 0 & 0 & 0 & 0 & 0 & 0 \\
20 &  ELR+WWLR+CLR & * & * & * & * & * & * & * & * & * & * & * & * & * & * & * & * & * & * & * & 0 & 0 & 0 & 0 \\
21 &  ELR+SWLR+CLR & * & * & * & * & * & * & * & * & * & * & * & * & * & * & * & * & * & * & * & 0 & 0 & 0 & 0 \\
22 &ELR+WWLR+CLR+TC & * & * & * & * & * & * & * & * & * & * & * & * & * & * & * & * & * & * & * & 0 & 0 & 0 & 0 \\
23 &ELR+SWLR+CLR+TC & * & * & * & * & * & * & * & * & * & * & * & * & * & * & * & * & * & * & * & * & * & 0 & 0 \\
\end{tabular}
\\
\\
\textbf{Head entities}\\

\begin{tabular}{llp{0.01cm}p{0.01cm}p{0.01cm}p{0.01cm}p{0.01cm}p{0.01cm}p{0.01cm}p{0.01cm}p{0.01cm}p{0.01cm}p{0.01cm}p{0.01cm}p{0.01cm}p{0.01cm}p{0.01cm}p{0.01cm}p{0.01cm}p{0.01cm}p{0.01cm}p{0.01cm}p{0.01cm}p{0.01cm}p{0.01cm}}
  &        Models  &01& 02& 03& 04& 05& 06& 07& 08& 09& 10& 11& 12& 13& 14& 15& 16& 17& 18& 19& 20& 21& 22& 23 \\
\hline
01 &           MFT & 0 & 0 & 0 & 0 & 0 & 0 & 0 & 0 & 0 & 0 & 0 & 0 & 0 & 0 & 0 & 0 & 0 & 0 & 0 & 0 & 0 & 0 & 0 \\
02 &  CLR(FORWARD) & * & 0 & 0 & 0 & 0 & 0 & 0 & 0 & 0 & 0 & 0 & 0 & 0 & 0 & 0 & 0 & 0 & 0 & 0 & 0 & 0 & 0 & 0 \\
03 &     CLR(LSTM) & * & * & 0 & 0 & 0 & 0 & * & 0 & 0 & 0 & 0 & 0 & 0 & 0 & 0 & 0 & 0 & 0 & 0 & 0 & 0 & 0 & 0 \\
04 &   CLR(BiLSTM) & * & * & 0 & 0 & 0 & 0 & * & 0 & 0 & 0 & 0 & 0 & 0 & 0 & 0 & 0 & 0 & 0 & 0 & 0 & 0 & 0 & 0 \\
05 &      CLR(CNN) & * & * & * & * & 0 & * & * & 0 & 0 & 0 & * & 0 & 0 & 0 & 0 & 0 & 0 & 0 & 0 & 0 & 0 & 0 & 0 \\
06 &      CLR(NSL) & * & * & * & * & 0 & 0 & * & 0 & 0 & 0 & 0 & 0 & 0 & 0 & 0 & 0 & 0 & 0 & 0 & 0 & 0 & 0 & 0 \\
07 &           BOW & * & * & 0 & 0 & 0 & 0 & 0 & 0 & 0 & 0 & 0 & 0 & 0 & 0 & 0 & 0 & 0 & 0 & 0 & 0 & 0 & 0 & 0 \\
08 &   WWLR(SkipG) & * & * & * & * & * & * & * & 0 & 0 & 0 & * & 0 & 0 & 0 & 0 & 0 & 0 & 0 & 0 & 0 & 0 & 0 & 0 \\
09 &  WWLR(SSkipG) & * & * & * & * & * & * & * & * & 0 & 0 & * & 0 & 0 & 0 & 0 & 0 & 0 & 0 & 0 & 0 & 0 & 0 & 0 \\
10 &          SWLR & * & * & * & * & * & * & * & 0 & 0 & 0 & * & 0 & 0 & 0 & 0 & 0 & 0 & 0 & 0 & 0 & 0 & 0 & 0 \\
11 &  BOW+CLR(NSL) & * & * & * & * & 0 & 0 & * & 0 & 0 & 0 & 0 & 0 & 0 & 0 & 0 & 0 & 0 & 0 & 0 & 0 & 0 & 0 & 0 \\
12 & WWLR+CLR(CNN) & * & * & * & * & * & * & * & * & 0 & * & * & 0 & * & 0 & 0 & 0 & 0 & 0 & 0 & 0 & 0 & 0 & 0 \\
13 & SWLR+CLR(CNN) & * & * & * & * & * & * & * & 0 & 0 & 0 & * & 0 & 0 & 0 & 0 & 0 & 0 & 0 & 0 & 0 & 0 & 0 & 0 \\
14 &    ELR(SkipG) & * & * & * & * & * & * & * & * & * & * & * & * & * & 0 & 0 & * & 0 & 0 & 0 & 0 & 0 & 0 & 0 \\
15 &   ELR(SSkipG) & * & * & * & * & * & * & * & * & * & * & * & * & * & 0 & 0 & * & 0 & 0 & 0 & 0 & 0 & 0 & 0 \\
16 &     AGG-FIGER & * & * & * & * & * & * & * & * & * & * & * & * & * & 0 & 0 & 0 & 0 & 0 & 0 & 0 & 0 & 0 & 0 \\
17 &       ELR+CLR & * & * & * & * & * & * & * & * & * & * & * & * & * & * & * & * & 0 & 0 & 0 & 0 & 0 & 0 & 0 \\
18 &      ELR+WWLR & * & * & * & * & * & * & * & * & * & * & * & * & * & * & * & * & 0 & 0 & 0 & 0 & 0 & 0 & 0 \\
19 &      ELR+SWLR & * & * & * & * & * & * & * & * & * & * & * & * & * & * & * & * & 0 & 0 & 0 & 0 & 0 & 0 & 0 \\
20 &  ELR+WWLR+CLR & * & * & * & * & * & * & * & * & * & * & * & * & * & * & * & * & 0 & 0 & 0 & 0 & 0 & 0 & 0 \\
21 &  ELR+SWLR+CLR & * & * & * & * & * & * & * & * & * & * & * & * & * & * & * & * & 0 & 0 & 0 & 0 & 0 & 0 & 0 \\
22 &ELR+WWLR+CLR+TC & * & * & * & * & * & * & * & * & * & * & * & * & * & * & * & * & * & 0 & 0 & 0 & 0 & 0 & 0 \\
23 &ELR+SWLR+CLR+TC & * & * & * & * & * & * & * & * & * & * & * & * & * & * & * & * & * & * & * & 0 & 0 & 0 & 0 \\
\hline
\end{tabular}\\
\\
\textbf{Tail entities}\\

\begin{tabular}{llp{0.01cm}p{0.01cm}p{0.01cm}p{0.01cm}p{0.01cm}p{0.01cm}p{0.01cm}p{0.01cm}p{0.01cm}p{0.01cm}p{0.01cm}p{0.01cm}p{0.01cm}p{0.01cm}p{0.01cm}p{0.01cm}p{0.01cm}p{0.01cm}p{0.01cm}p{0.01cm}p{0.01cm}p{0.01cm}p{0.01cm}}
  &        Models  & 01& 02& 03& 04& 05& 06& 07& 08& 09& 10& 11& 12& 13& 14& 15& 16& 17& 18& 19& 20& 21& 22& 23 \\
\hline
01 &           MFT & 0 & 0 & 0 & 0 & 0 & 0 & 0 & 0 & 0 & 0 & 0 & 0 & 0 & 0 & 0 & 0 & 0 & 0 & 0 & 0 & 0 & 0 & 0 \\
02 &  CLR(FORWARD) & * & 0 & 0 & 0 & 0 & 0 & 0 & 0 & 0 & 0 & 0 & 0 & 0 & 0 & 0 & 0 & 0 & 0 & 0 & 0 & 0 & 0 & 0 \\
03 &     CLR(LSTM) & * & * & 0 & 0 & 0 & 0 & 0 & 0 & 0 & 0 & 0 & 0 & 0 & 0 & 0 & 0 & 0 & 0 & 0 & 0 & 0 & 0 & 0 \\
04 &   CLR(BiLSTM) & * & * & * & 0 & 0 & 0 & * & 0 & 0 & 0 & 0 & 0 & 0 & 0 & 0 & 0 & 0 & 0 & 0 & 0 & 0 & 0 & 0 \\
05 &      CLR(CNN) & * & * & * & * & 0 & * & * & * & 0 & 0 & * & 0 & 0 & 0 & 0 & 0 & 0 & 0 & 0 & 0 & 0 & 0 & 0 \\
06 &      CLR(NSL) & * & * & * & * & 0 & 0 & * & 0 & 0 & 0 & * & 0 & 0 & 0 & 0 & 0 & 0 & 0 & 0 & 0 & 0 & 0 & 0 \\
07 &           BOW & * & * & 0 & 0 & 0 & 0 & 0 & 0 & 0 & 0 & 0 & 0 & 0 & 0 & 0 & 0 & 0 & 0 & 0 & 0 & 0 & 0 & 0 \\
08 &   WWLR(SkipG) & * & * & * & * & 0 & 0 & * & 0 & 0 & 0 & * & 0 & 0 & 0 & 0 & 0 & 0 & 0 & 0 & 0 & 0 & 0 & 0 \\
09 &  WWLR(SSkipG) & * & * & * & * & 0 & 0 & * & 0 & 0 & 0 & * & 0 & 0 & 0 & 0 & 0 & 0 & 0 & 0 & 0 & 0 & 0 & 0 \\
10 &          SWLR & * & * & * & * & * & * & * & * & * & 0 & * & 0 & 0 & 0 & 0 & 0 & 0 & 0 & 0 & 0 & 0 & 0 & 0 \\
11 &  BOW+CLR(NSL) & * & * & * & * & 0 & 0 & * & 0 & 0 & 0 & 0 & 0 & 0 & 0 & 0 & 0 & 0 & 0 & 0 & 0 & 0 & 0 & 0 \\
12 & WWLR+CLR(CNN) & * & * & * & * & * & * & * & * & * & * & * & 0 & 0 & 0 & 0 & * & 0 & 0 & 0 & 0 & 0 & 0 & 0 \\
13 & SWLR+CLR(CNN) & * & * & * & * & * & * & * & * & * & * & * & 0 & 0 & 0 & 0 & 0 & 0 & 0 & 0 & 0 & 0 & 0 & 0 \\
14 &    ELR(SkipG) & * & * & * & * & * & * & * & * & * & * & * & * & * & 0 & 0 & * & 0 & 0 & 0 & 0 & 0 & 0 & 0 \\
15 &   ELR(SSkipG) & * & * & * & * & * & * & * & * & * & * & * & * & * & * & 0 & * & 0 & 0 & 0 & 0 & 0 & 0 & 0 \\
16 &     AGG-FIGER & * & * & * & * & * & * & * & * & * & 0 & * & 0 & 0 & 0 & 0 & 0 & 0 & 0 & 0 & 0 & 0 & 0 & 0 \\
17 &       ELR+CLR & * & * & * & * & * & * & * & * & * & * & * & * & * & * & * & * & 0 & 0 & 0 & 0 & 0 & 0 & 0 \\
18 &      ELR+WWLR & * & * & * & * & * & * & * & * & * & * & * & * & * & * & * & * & 0 & 0 & 0 & 0 & 0 & 0 & 0 \\
19 &      ELR+SWLR & * & * & * & * & * & * & * & * & * & * & * & * & * & * & * & * & 0 & 0 & 0 & 0 & 0 & 0 & 0 \\
20 &  ELR+WWLR+CLR & * & * & * & * & * & * & * & * & * & * & * & * & * & * & * & * & * & 0 & 0 & 0 & 0 & 0 & 0 \\
21 &  ELR+SWLR+CLR & * & * & * & * & * & * & * & * & * & * & * & * & * & * & * & * & * & * & * & 0 & 0 & 0 & 0 \\
22 &ELR+WWLR+CLR+TC & * & * & * & * & * & * & * & * & * & * & * & * & * & * & * & * & * & * & * & 0 & 0 & 0 & 0 \\
23 &ELR+SWLR+CLR+TC & * & * & * & * & * & * & * & * & * & * & * & * & * & * & * & * & * & * & * & * & 0 & 0 & 0 \\
\end{tabular}

\end{tabular}

\caption{Significance-test results for accuracy measure
for all, head and tail entities.
If the result for the model in a row is significantly larger than the result for the model 
in a column, then the value in the corresponding (row,column) is *
and otherwise is 0.
}
\tablabel{sigTable}
\end{center}
\end{table*}

\end{document}